\documentclass[letterpaper, 10 pt, conference]{ieeeconf}  

\usepackage{amsmath}
\usepackage{hyperref}
\IEEEoverridecommandlockouts                              

\usepackage{graphicx}
\usepackage{algorithm}
\usepackage{algorithmic}
\usepackage{mathtools}
\usepackage{xcolor}
\usepackage{subfig}

\title{\LARGE \bf
Robust Humanoid Locomotion Using Trajectory Optimization and Sample-Efficient Learning*}

\author{Mohammad Hasan Yeganegi$^{1}$, Majid Khadiv$^{2}$, S. Ali A. Moosavian$^{1}$,\\ Jia-Jie Zhu$^{2}$, Andrea Del Prete$^{3}$, and Ludovic Righetti$^{2,4}$
\thanks{*This work is supported by New York University, the Max-Planck Society, the European Union’s Horizon 2020 research and innovation program (grant agreement No 780684 and European Research Council’s grant No 637935) and the National Science Foundation (grant CMMI-1825993)}
\thanks{Jia-Jie Zhu is supported by the European Unions Horizon 2020 research and innovation programme under the Marie Sk\l{}odowska-Curie grant agreement No 798321.}
\thanks{$^{1}$ K. N. Toosi University of Technology, Tehran, Iran.
        {\tt\small yeganegi.m.h@email.kntu.ac.ir}
        {\tt\small moosavian@kntu.ac.ir}}%
\thanks{$^{2}$ Max Planck Institute for Intelligent Systems, Tuebingen, Germany. {\tt\small firstname.lastname@tuebingen.mpg.de}}%
\thanks{$^{3}$ Industrial Engineering Department, University of
Trento, Italy. {\tt\small andrea.delprete@unitn.it}}
\thanks{$^{4}$ Tandon School of Engineering, New York University, New York, USA. {\tt\small ludovic.righetti@nyu.edu}}%
}

\begin{document}
\maketitle
\thispagestyle{empty}
\pagestyle{empty}

\begin{abstract}

Trajectory optimization (TO) is one of the most powerful tools for generating feasible motions for humanoid robots. However, including uncertainties and stochasticity in the TO problem to generate robust motions can easily lead to intractable problems. Furthermore, since the models used in TO have always some level of abstraction, it can be hard to find a realistic set of uncertainties in the model space. In this paper we leverage a sample-efficient learning technique (Bayesian optimization) to robustify TO for humanoid locomotion. The main idea is to use data from full-body simulations to make the TO stage robust by tuning the cost weights. To this end, we split the TO problem into two phases. The first phase solves a convex optimization problem for generating center of mass (CoM) trajectories based on simplified linear dynamics. The second stage employs iterative Linear-Quadratic Gaussian (iLQG) as a whole-body controller to generate full body control inputs. Then we use Bayesian optimization to find the cost weights to use in the first stage that yields robust performance in the simulation/experiment, in the presence of different disturbance/uncertainties. The results show that the proposed approach is able to generate robust motions for different sets of disturbances and uncertainties.

\end{abstract}

\section{INTRODUCTION}
Given the high complexity of humanoid robots (due to redundancy, underactuation, hybrid dynamics), generating feasible and optimal motions for them is extremely challenging. Trajectory optimization (TO) is among the best tools available to take into account all the physical and geometrical constraints, while at the same time generating locally optimal dynamic motions. However, to be able to solve the TO problem quickly, we often resort to using simplified models of the robot (e.g. linear inverted pendulum model or centroidal momentum dynamics). In this case, the discrepancy between the model and the real robot, as well as uncertainties in the constraint sets (center of pressure (CoP), friction cone, etc.), could cause brittle or infeasible motions. One way to overcome this issue is simply to use conservative constraint sets (such as shrinking the support polygon for a CoP constraint), but this limits the capability of TO to find desired aggresive solutions. The other problem is that finding realistic values for the margins of the constraints is challenging and needs trial and error.  

One principled way to deal with this problem is to use stochastic or robust trajectory optimization approaches to explicitly add model uncertainties to the problem. Although systematic, this approach suffers from two main issues in humanoid locomotion. First, identifying different types of uncertainties and projecting them to a realistic set of constraints in the simplified model space is challenging and not always feasible. Second, adding stochastic uncertainties to the problem can easily lead to an intractable problem, and in most cases it can be solved only for simplified worst-case scenarios, where it tends to be very conservative. 

\begin{figure}[!t]
    \centering
    \includegraphics[clip,trim=5cm 4.6cm 7cm 4.7cm,width=.48\textwidth]{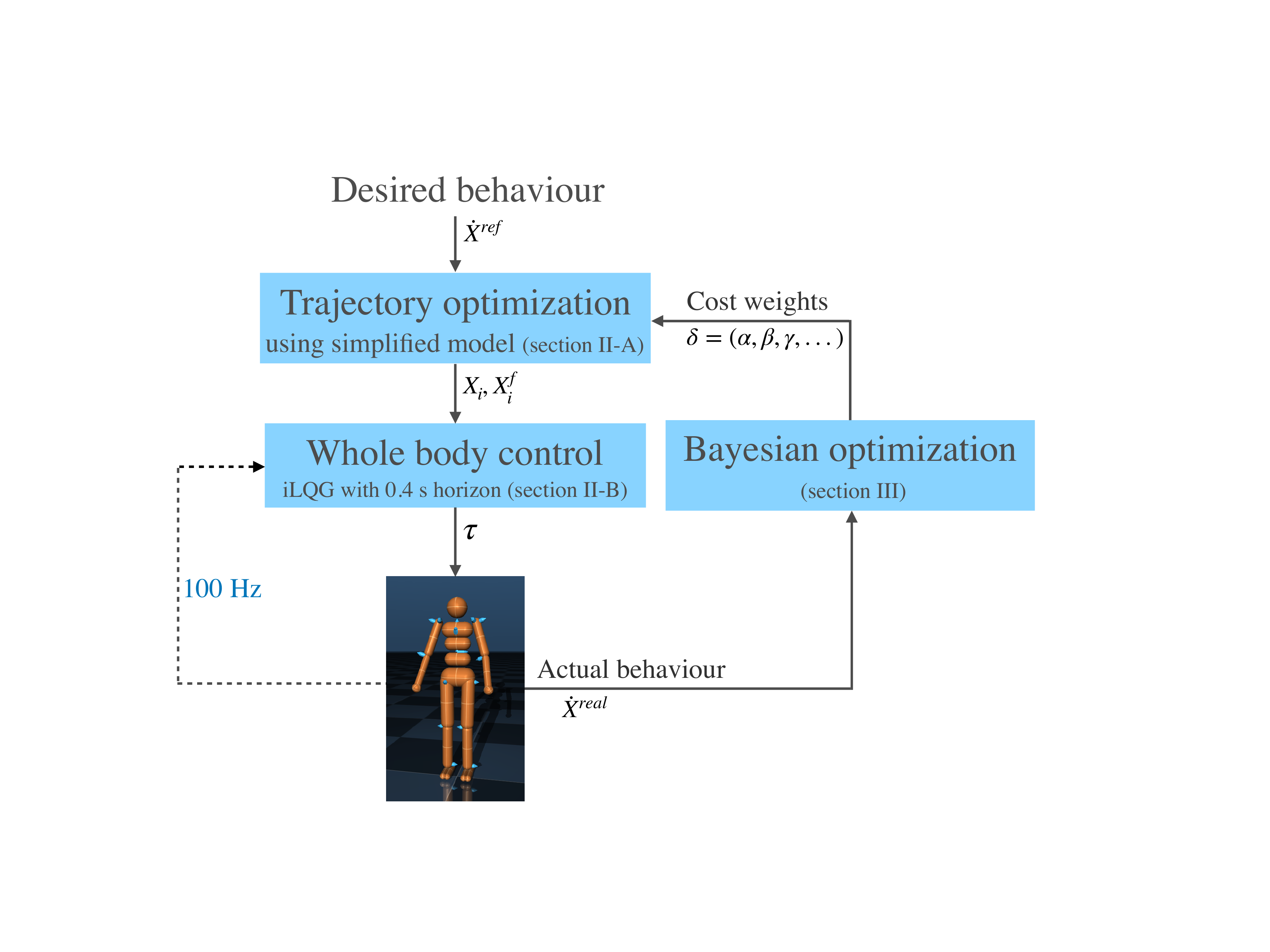}
    \caption{A high-level block diagram of the proposed approach}
    \label{fig:block_diagram}
    \vspace{-5 mm}
\end{figure}

In this paper, we combine the strengths of trajectory optimization and sample-efficient learning (data-driven black box optimization) to generate robust motions for different kinds of uncertainties with a low number of experiments. The main idea is to use data from full-body simulations to increase the robustness of the TO stage (Fig. \ref{fig:block_diagram}). We formulate the TO problem such that changing the cost function weights trades off robustness against performance. Then, we employ Bayesian optimization (BO) to find the cost weights of TO that achieve the task at best in the presence of disturbances and uncertainties.  

\subsection{Trajectory optimization for humanoid locomotion}
Starting from the seminal work of Kajita et al.~\cite{kajita2003biped}, TO has become the dominant approach to generate motions for humanoid robots. This work uses the linear inverted pendulum model (LIPM) \cite{kajita20013d} and an infinite horizon linear quadratic program (named preview control) to generate center of mass (CoM) trajectories given the predefined footstep locations. This work has been extended in different ways, adding inequality constraints on the ZMP \cite{wieber2006trajectory}, adaptive footstep locations \cite{herdt2010online}, and friction cone constraints~\cite{khadiv2017pattern}. However, this model can only describe the underlying dynamics of walking on flat terrain with zero angular momentum around the CoM.

For more complex motions such as walking on uneven terrains and multi-contact scenarios, researchers use the centroidal momentum dynamics~\cite{orin2013centroidal}. Although this model describes the unactuated part of the robot dynamics exactly \cite{wieber2016modeling}, it is nonlinear and renders the corresponding optimization problem non-convex~\cite{herzog2015trajectory,carpentier2016A,ponton2018time}. At the same time, in the centroidal dynamics model, constraints on the state (CoM position, velocity and angular momentum) as well as the control (centroidal wrench) are a function of full body motion and joint torque constraints. In fact, this problem is inherent in the use of any simplified model.

One systematic way to find this constraint set is to learn the feasible set in the centroidal space from extensive full-body simulations~\cite{carpentier2017learning}. However, this approach needs too many simulations for learning all the constraints and it does not cover all the uncertainties in a real simulation or experiment (e.g. uncertainty in contact locations and timings, or change in the environment). Even if one could use the full body model of a humanoid to take all the robot constraints into account exactly in a computationally efficient manner \cite{lengagne2013generation} (which makes the problem high-dimensional and non-convex), 
the models will always fail to capture perfectly reality and uncertainty in the models of robot and environment can still make the solution brittle.

\subsection{Bayesian optimization for humanoid locomotion}
Bayesian optimization is a form of black-box optimization. Black-box and derivative-free optimization has a long history in numerical optimization as well as in statistics \cite{shahriari2016taking}. In robotics, BO has been applied to different problems, e.g. locomotion of a quadruped parameterized by a walk engine \cite{lizotte2007automatic}, locomotion of a hexapod \cite{cully2015robots}, balancing of an inverted pendulum \cite{marco2016automatic}, finding task priority in inverse dynamics for controlling dual-arm manipulators \cite{su2018sample}, and  scheduling contact for a one-leg hopper \cite{seyde2019locomotion}. For example, \cite{marco2016automatic} applied BO to tune performance cost of an unconstrained infinite-horizon LQR for controlling an inverted pendulum on an experimental setting. Contrary to \cite{marco2016automatic}, here we employ BO to tune the cost of a constrained trajectory optimization for a humanoid robot where the cost trades off performance against robustness.

Application of BO to humanoid locomotion is limited to simplified cases such as planar bipeds \cite{calandra2014experimental,antonova2016sample,rai2018bayesian}. In \cite{calandra2014experimental}, the authors employed BO to learn a parametrized walking policy of a small-sized planar biped robot optimizing eight parameters:  four control signals to the knee and hip joints as well as four parameters of a finite state machine. \cite{antonova2016sample} applied BO to a more complex model of a planar biped robot, where they used 16 neuromuscular policy variables to parameterize walking of a planar biped robot. The parameterization comprises of 10 variables for stance phase feedback gains and 6 parameters for swing phase. They applied their approach to a simulation of a 7-link robot walking on uneven as well as sloped surfaces. They applied a generalized version of this approach on the simulation and experiment of the biped robot ATRIAS \cite{rai2018bayesian}. Both \cite{calandra2014experimental} and \cite{antonova2016sample} mentioned that only a very low percentage of the parameter space leads to a feasible gait, which shows the difficulty of generating feasible motions for humanoid robots using \textit{only} black-box optimization. Contrary to these works, we resort to using constrained gradient-based trajectory optimization to generate feasible trajectories and use Bayesian optimization on top of it to make the trajectories robust by automatically tuning the cost weights. Also, \cite{charbonneau2018learning,yuan2019bayesian} used BO to find the parameters of a whole-body controller for a humanoid robot, yielding robust performance for the control. Our work can be seen as complementary to these work, because we propose to use BO to find the best cost weights of TO for a given whole-body controller.

\subsection{Contribution}
The main contribution of this paper is to propose a framework that combines gradient-based and gradient-free optimization for generating robust humanoid locomotion. This framework uses a simplified model and deterministic proxy constraints for the TO problem, where the cost terms trade off performance against robustness. Writing the problem in this way, we could have different cost weights that push solutions away from the boundaries of the constraint sets. Then we use the full-body simulation of a humanoid robot (with disturbances and uncertainties) and exploit BO to efficiently find a set of TO cost weights that achieve the task at best while satisfying the full robot constraints. Fig. \ref{fig:block_diagram} shows a block diagram of the proposed framework.


\section{Optimal control problem}\label{section:OC}
\subsection{First Stage : Convex Trajectory Optimization for Walking}
This section describes the TO approach~\cite{khadiv2017pattern} that we use for generating CoM trajectories given a desired walking velocity. Note however that in principle any other algorithm could be used. In~\cite{khadiv2017pattern} walking is formulated as a trade off between three cost terms: desired velocity tracking, foot tip-over avoidance, and slippage avoidance. 
\begin{align}
    \label{eq:TO}
    \underset{\dddot X_i , X_i^f}{\text{min.}} \, \sum\limits_{i=1}^{N}\, & \alpha \Vert \dot X_{i}-\dot X_{i}^{ref} \Vert^2  + \beta \Vert Z_{i}-Z_{i}^{ref} \Vert^2  + \gamma \Vert \mu_{i} \Vert^2  \nonumber\\
    \text{s.t.} \qquad &\mu_{i} \in friction\,cone \quad , \quad \forall i=1,...,N. \nonumber\\
    \qquad &X_i^f \in reachable\,area \quad , \quad \forall i=1,...,N. \nonumber\\
     \qquad &Z_{i} \in support\,polygon \quad , \quad \forall i=1,...,N.
\end{align}
where $X=[c_x,c_y]^T$ is the horizontal CoM position. $Z=[z_x,z_y]^T$ is the zero moment point (ZMP) position and $\mu$ is the required coefficient of friction (RCoF). $\dot{X}^{ref}$ is the desired walking velocity, $Z^{ref}$ is the desired ZMP, which is taken at the center of the foot to maximize the feasibility margins. As shown in \cite{khadiv2017pattern}, \eqref{eq:TO} can be written as a quadratic program (QP), assuming the linear inverted pendulum dynamics and polyhedral approximation of friction cones. This program yields consistent CoM trajectory and foot locations for a given desired walking velocity. Then, we use polynomials to generate the swing foot trajectories.

Depending on the cost weights $\alpha, \beta, \gamma$, we get different CoM trajectories. For example, if $\beta=\gamma=0$ the optimizer generates a feasible CoM motion, while trying to achieve the desired walking velocity. However, ZMP and RCoF might reach their boundaries. As a result, even if the whole-body controller can track this CoM trajectory, a slight disturbance could cause a fall (or infeasibility in MPC setting). On the other hand, with high values of $\beta, \gamma$ the optimizer generates CoM trajectories and foot locations leading to high margins for ZMP and RCoF, at the expenses of the velocity tracking. As a result, it is crucial to find the optimal values of these weights, which generate enough constraint margins while achieving the task at best.

\subsection{Second Stage: iLQG for generating whole-body torques}
In this section, we use an iterative linear quadratic Gaussian (iLQG) controller to map the desired CoM and feet trajectories from the first stage to the whole-body torques, while penalizing the full-body constraints~\cite{tassa2012synthesis}. iLQG linearizes the dynamics and computes a second-order approximation of the cost around a nominal trajectory. In the backward pass, feedforward and feedback terms are obtained, accounting for box constraints on the control~\cite{tassa2014control}. Finally, convergence of the cost is achieved by applying a line search~\cite{tassa2012synthesis}. Note that we use iLQG as a whole-body controller with a short horizon of 0.4 s to track the desired trajectories from the first stage. The humanoid robot has 27 DoFs, it is 1.37 m tall and weighs 41 kg. Abdomen, shoulder and ankle joints are 2-DoF, while elbows, knees and pelvis are 1-DoF and hips are 3-DoF joints. 

The cost function in our problem is comprised of the following terms~\cite{erez2013integrated}:
\begin{itemize}
    \item Quadratic costs of feet and COM velocity tracking errors.
    \item Smooth-abs” function \cite{tassa2012synthesis} to track the desired feet and COM positions.
    \item  Quadratic costs to minimize joint torques, joint velocities, angular velocity of the pelvis, linear velocity of torso in vertical direction, and finally angular velocity of the feet around vertical  direction.
    \item Quadratic costs to penalize the deviations between the orientation of the pelvis, torso and the two feet.
    \item Quadratic costs to penalize deviations of the Z axis of torso and both feet from the global vertical direction.
    \item Quadratic costs to penalize deviation of the global height of the torso from the fixed value used for the LIPM.
\end{itemize}

\section{Hyper-parameter tuning via bayesian optimization} \label{section:BO}
\subsection{Problem formulation for robust humanoid locomotion}
We propose to close the loop in our system, i.e., we formulate an overall optimization problem based on the quantities in Fig. \ref{fig:block_diagram}:
\begin{align}
\label{eq:opt_final}
\underset{\delta}{\text{min.}\nonumber} \quad & J(\delta)\vcentcolon = \sum\limits_{i=1}^{N}\, \Vert \dot X^{\text{real}}_{i}(\delta) -\dot X_{i}^{des} \Vert^2 + \lambda \phi(h_N^\delta),\\
\text{s.t.} \quad & \dot X^{\text{real}}_{i}(\delta)\quad \text{is the output of simulation in Fig~\ref{fig:block_diagram}.}
\end{align}
$\delta=(\alpha, \beta, \gamma)$ is the collection of the hyper-parameters used in optimization problem \eqref{eq:TO}. $\dot X^{\text{real}}_{i}(\delta)$ is the CoM velocity obtained by solving the QP \eqref{eq:TO} and applying iLQG tracking to the simulation of the robot full body with different (unknown) disturbances. 
$h^\delta_N$ is the CoM height at the final time step, $\lambda$ is a user-defined weight, $\phi(.)$ is a function used to penalize falls, e.g. $\phi = \max (|h^\delta_N-h_{des}|-\text{threshold}, 0)$.
In the next subsection, we detail how the optimization problem~\eqref{eq:opt_final} is solved using BO. 

\begin{figure}[!t]
    \centering
    \subfloat{\includegraphics[width=1.4cm]{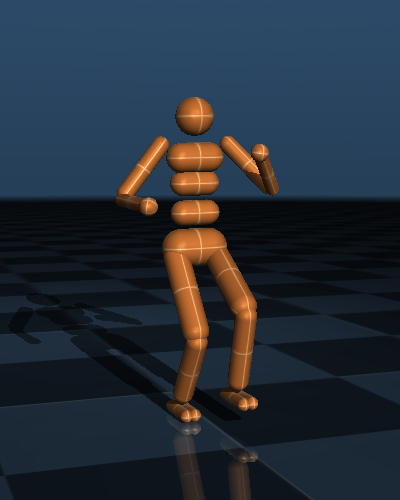}}
    \subfloat{\includegraphics[width=1.4cm]{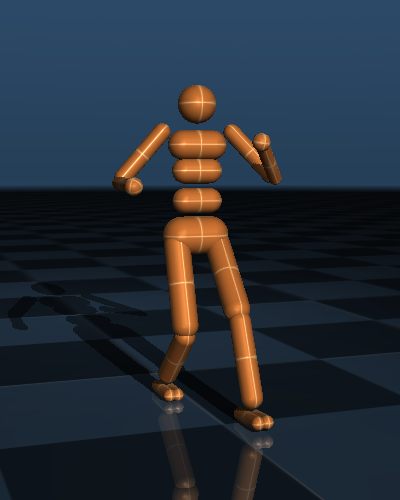}}
    \subfloat{\includegraphics[width=1.4cm]{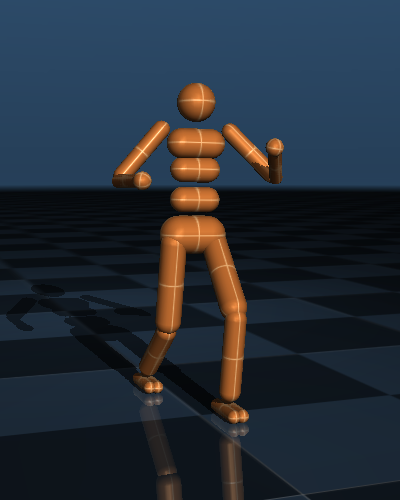}}
    \subfloat{\includegraphics[width=1.4cm]{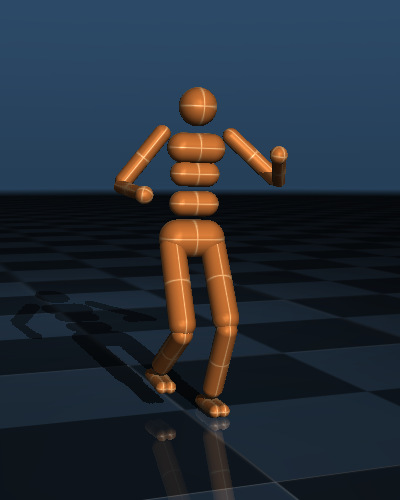}}
    \subfloat{\includegraphics[width=1.4cm]{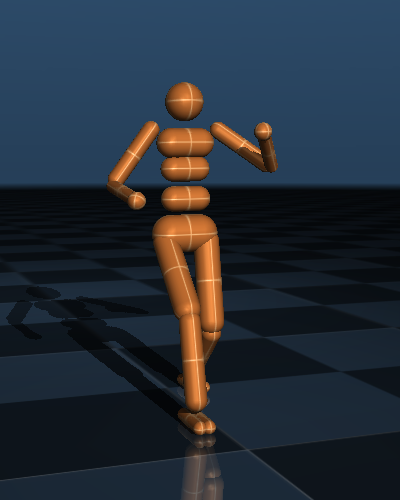}}
    \subfloat{\includegraphics[width=1.4cm]{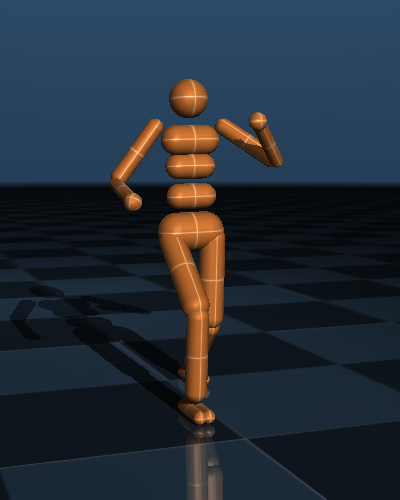}}\\
    \subfloat{\includegraphics[width=1.4cm]{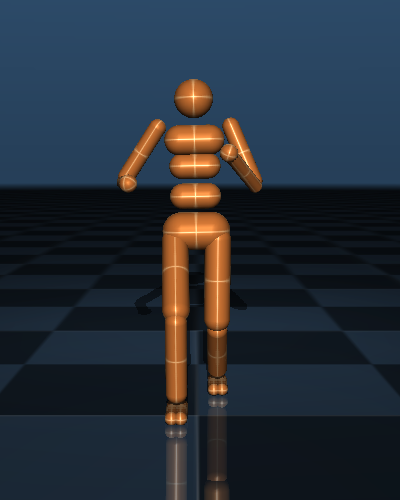}}
    \subfloat{\includegraphics[width=1.4cm]{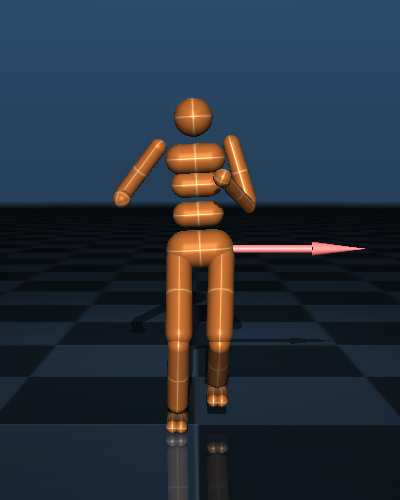}}
    \subfloat{\includegraphics[width=1.4cm]{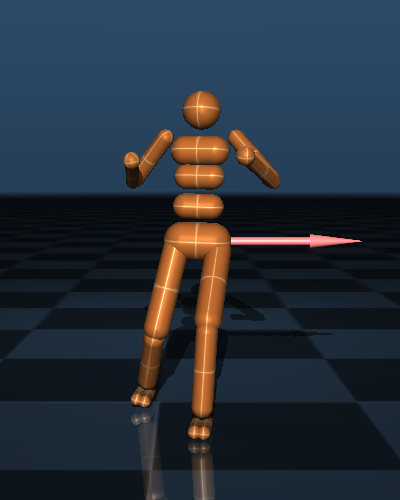}}
    \subfloat{\includegraphics[width=1.4cm]{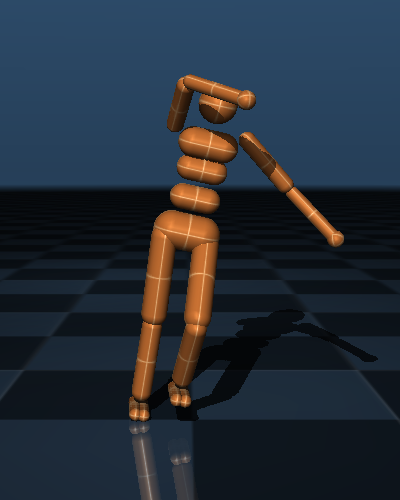}}
    \subfloat{\includegraphics[width=1.4cm]{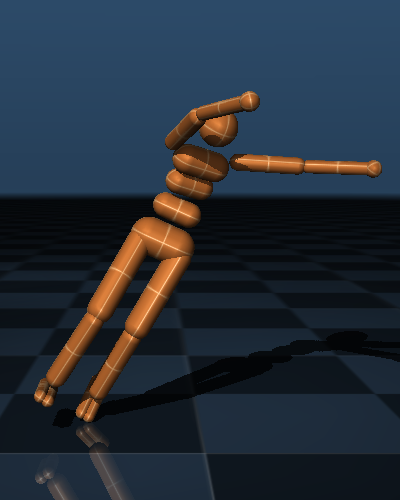}}
    \subfloat{\includegraphics[width=1.4cm]{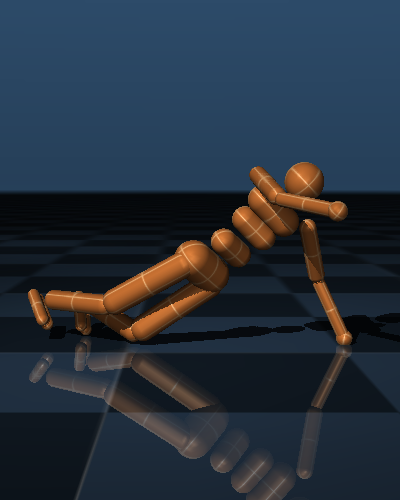}}
    \caption{Screenshots of simulation of a 27 DoF humanoid robot, (top) without disturbance (bottom) with lateral push}
    \label{screenshot}
    \vspace{-5 mm}
\end{figure}

\subsection{BO algorithm}
We apply BO techniques that make use of Gaussian processes (GP) to model the unknown process -- the objective function $J$ in \eqref{eq:opt_final}. 
The distribution of GP naturally contains information about the uncertainty of $J$.
Intuitively, BO trades off exploration (high-variance) and exploitation (high-value).
This is achieved by maximizing an acquisition function that captures this trade-off.
An example of the acquisition function is the upper (or lower)-confidence bound (UCB) (cf. e.g. \cite{brochu2010tutorial})
\begin{align}
\label{eq:ucb}
u_{\text{UCB}}(x)=\mu(x) + \kappa \sigma(x),
\end{align}
where $\mu(x)$ and $\sigma(x)$, mean and standard deviation, are computed by the current estimates of the Gaussian process distribution. Intuitively, in the beginning, the algorithm explores where $\sigma(x)$ is high. As we collected more data, $\sigma(x)$ decreases and the algorithm exploits good regions of the objective function with high $\mu(x)$. We illustrate this effect in Fig.~\ref{weights}.

For the BO problem we have used scikit-optimize\footnote{\url{https://github.com/scikit-optimize/scikit-optimize}}. We used $gp-hedge$ as acquisition function, which is a probabilistic combination of the UCB \eqref{eq:ucb}, expected improvement and probability of improvement \cite{hoffman2011portfolio}. 

Algorithm~\ref{alg:bo} shows how problem~\eqref{eq:opt_final} is solved in this setting.
\begin{algorithm}[tbp]
	\caption{Pseudo code for Bayesian optimization}
	\label{alg:bo}
	\begin{algorithmic}[1]
	    \STATE Given: A black-box function $J$ for evaluation (without analytical gradient), an acquisition function $u$ (e.g. $u_\text{UCB}$ in \eqref{eq:ucb}).
	    \STATE Output: Current best minimizer $\delta^*$ of $J(\delta)$
	    \STATE Initialize with a dataset $\mathcal{D}=\{(\delta_i, y_i)\}_{i=1, 2, \dots}$, best objective value $y_\text{best}=\max_i{y_i}$
		\REPEAT
        \STATE Find next query parameter $\delta_t$ by maximizing the acquisition function $u$
        $$\delta_t = \arg\max_{\delta}{u(\delta|\mathcal{D})}.$$ This is carried out by a numerical optimization routine, e.g. L-BFGS.
        \STATE Evaluate the objective function at $\delta_t$
        $$y_t = J(\delta_t).$$\\
        If $y_t < y_\text{best}$, set $\delta_\text{best} \gets \delta_t, \, y_\text{best} \gets y_t$
        \STATE  Add the new data point to the dataset 
        $$\mathcal{D} \gets\mathcal{D}\cup (\delta_t, y_t).$$ 
        Update $\mathcal{GP}(\mu(\delta), \sigma(\delta))$ and the resulting acquisition function $u(\delta|\mathcal{D})$.
        \UNTIL Computation budget reached.
	\end{algorithmic}
\end{algorithm}
It may be helpful to think of BO as using GP as a surrogate for the unknown objective
$$
y := J(\delta) \sim \mathcal{GP}(\mu(\delta), \sigma(\delta)),
$$
where $\mu(\delta), \sigma(\delta)$ are respectively the mean and standard deviation of the GP distribution evaluated at point $\delta$.
As we obtain more data, the GP approximates the objective better.

Given the dataset $\mathcal{D}=\{\delta_i, y_i\}_{i=1, 2, \dots}$, the GP distribution at a new point $\delta_*$ is computed using the formula:
\begin{align}
\label{eq:gp}
\mu(\delta_*) &= K_*^{T} K^{-1} y\nonumber\\
\sigma(\delta_*) &= K_{**} - K_*^{T} K^{-1} K_*,
\end{align}
where $K_*=k(\delta_*, \boldsymbol{\delta}), \, K=k(\boldsymbol{\delta}, \boldsymbol{\delta}), K_{**}=k(\delta_*, \delta_*).$ 
The notation $\boldsymbol{\delta}$ denotes the vector of parameters already stored in the data set, i.e., $\boldsymbol{\delta} = (\delta_1, \delta_2, \dots)$.
$k(x,x')$ is a kernel that measures the similarity between $x$ and $x'$. For example, it is computed in one-dimension by
$$
k(x,x') = a \, \text{exp} \left (-\frac{1}{2b}(x-x')^2 \right ),
$$
where $a,b$ are kernel parameters.

\section{Results}\label{section:results}
To show the effectiveness of the proposed framework, we present three scenarios in this section. 
In the first scenario, we present a practical example that shows how the choice of cost in the TO problem affects robustness and performance. Then, in the second scenario we apply BO to the TO problem with two cost weights and analyse how BO converges to the optimal set of cost weights. Finally in the third scenario we show how our approach can scale to cost functions with a larger number of weights.\footnote{A summary of our humanoid simulations on different scenarios is available at: \url{https://www.youtube.com/watch?v=iek_goPaF9w&feature=youtu.be}}

\subsection{Scenario 1: Walking with different cost weights}
\begin{figure}[!t]
    \centering
    \subfloat[Walking without disturbance]{\includegraphics[clip,trim=0.2cm .3cm 0.2cm .2cm,width=.45\textwidth]{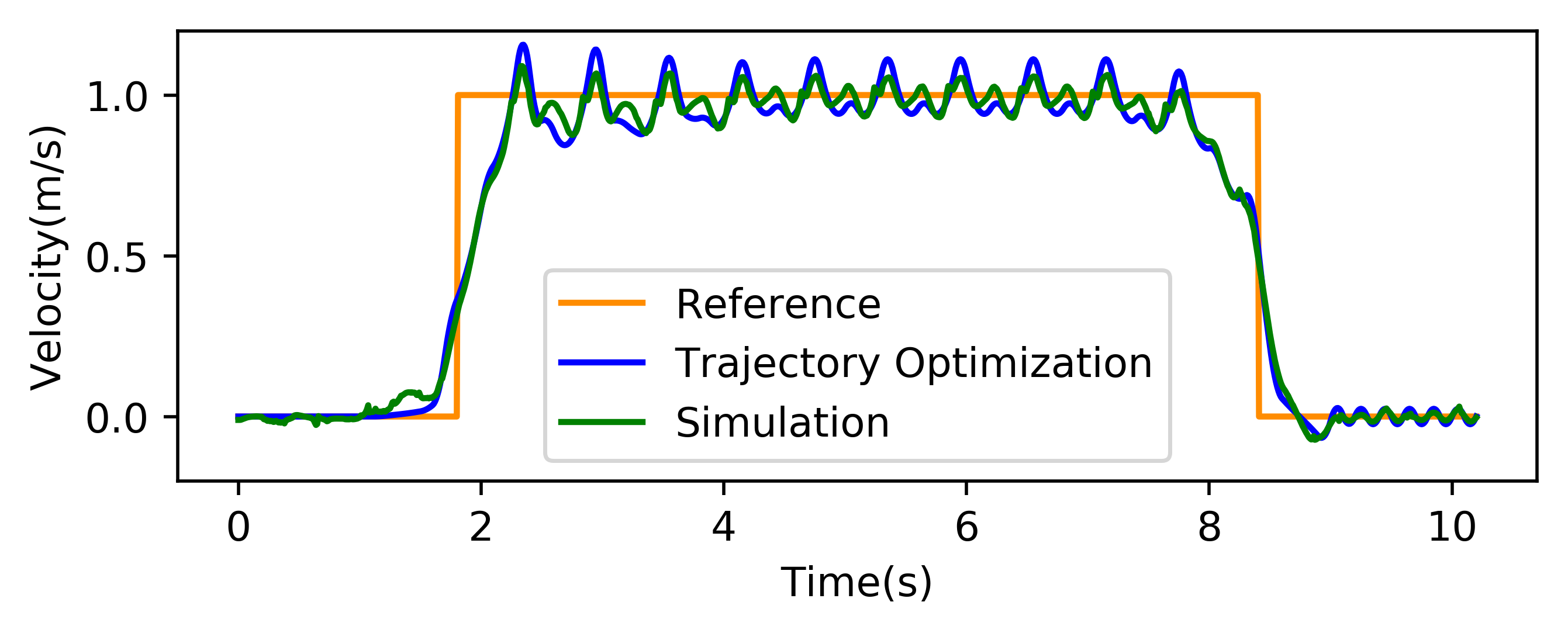}\label{xvel_com_vel_tracking}}\\
    \subfloat[Walking with external push]{\includegraphics[clip,trim=0.2cm .3cm 0.2cm .2cm,width=.45\textwidth]{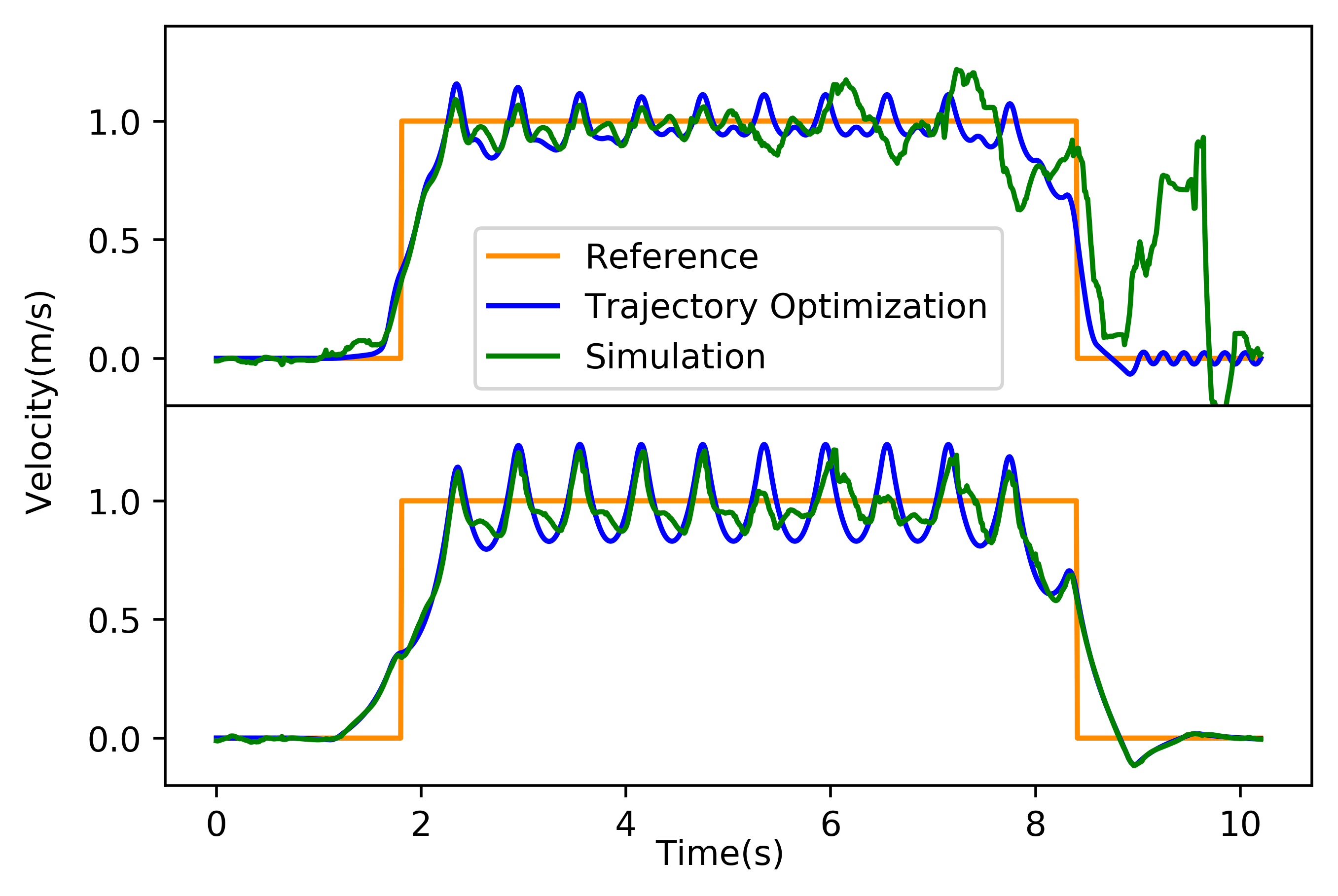}\label{xvel_com_push}}\\
    \subfloat[Walking on slippery terrain ]{\includegraphics[clip,trim=0.2cm .3cm 0.2cm .2cm,width=.45\textwidth]{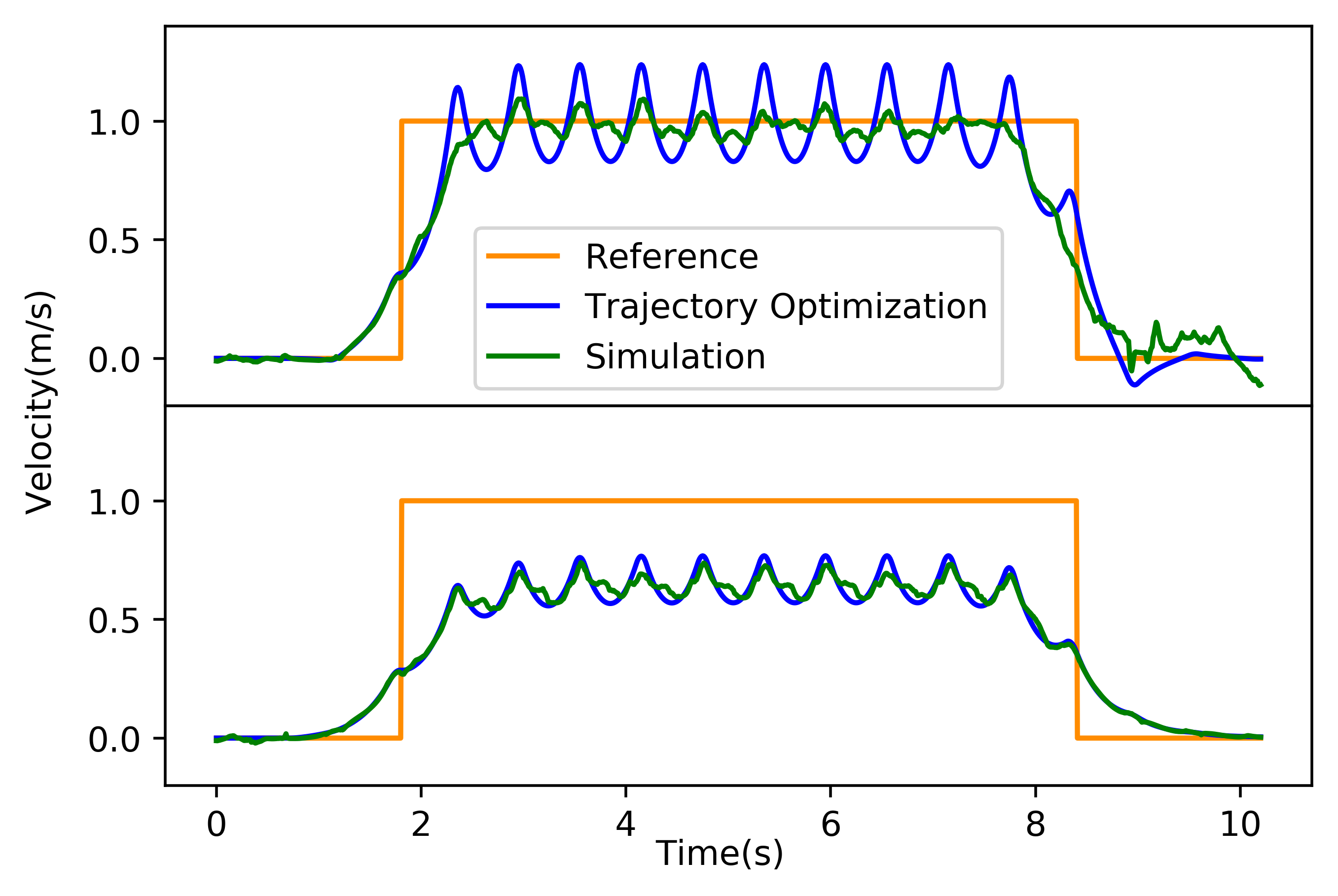}\label{xvel_com_slippage}}
    \caption{
    Scenario 1: velocity tracking \protect\subref{xvel_com_vel_tracking} with the set of weights $\alpha=1$ and $\beta=0, \gamma=0$, the robot succeeds to finish the walking tracking the velocity well. \protect\subref{xvel_com_push} the robot is pushed during walking (see the supplementary video), with the set of weights in previous case it falls down (top). However, increasing $\beta$ to 70 makes the trajectory robust and the robot is able to finish the task successfully (bottom). \protect\subref{xvel_com_slippage} Decreasing the friction coefficient in the simulation to 0.15 while the friction cone limit in TO is 0.4, the robot falls down (see the supplementary video) with the previous weights (top), increasing $\gamma$ to 30 enables the robot to finish the task at the cost of velocity tracking degradation (bottom). }
    \label{xvel_com}
    \vspace{-5 mm}
\end{figure}

\begin{figure*}[!t]
    \centering
    \subfloat[Walking without disturbance]{\includegraphics[clip,trim=0.2cm .3cm 0.2cm .2cm,width=.95\textwidth]{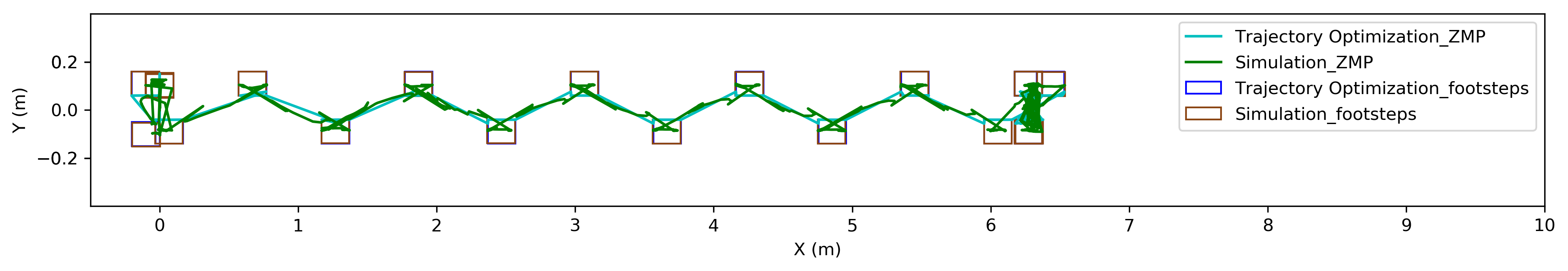}\label{footstep_vel_tracking}}\\
    \subfloat[Walking with external push]{\includegraphics[clip,trim=0.2cm .3cm 0.2cm .2cm,width=.95\textwidth]{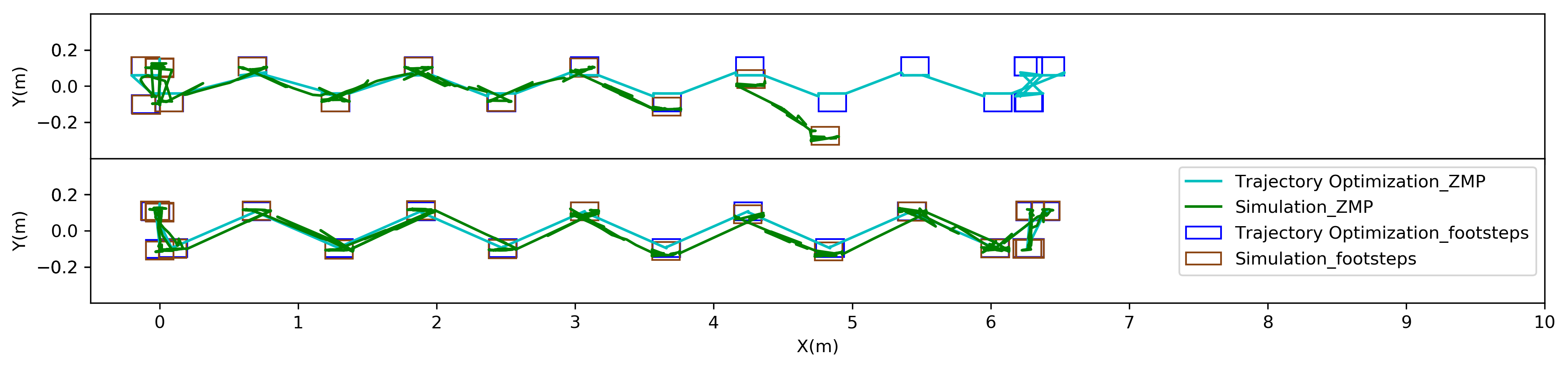}\label{footstep_push}}\\
    \subfloat[Walking on slippery terrain ]{\includegraphics[clip,trim=0.2cm .3cm 0.2cm .2cm,width=.95\textwidth]{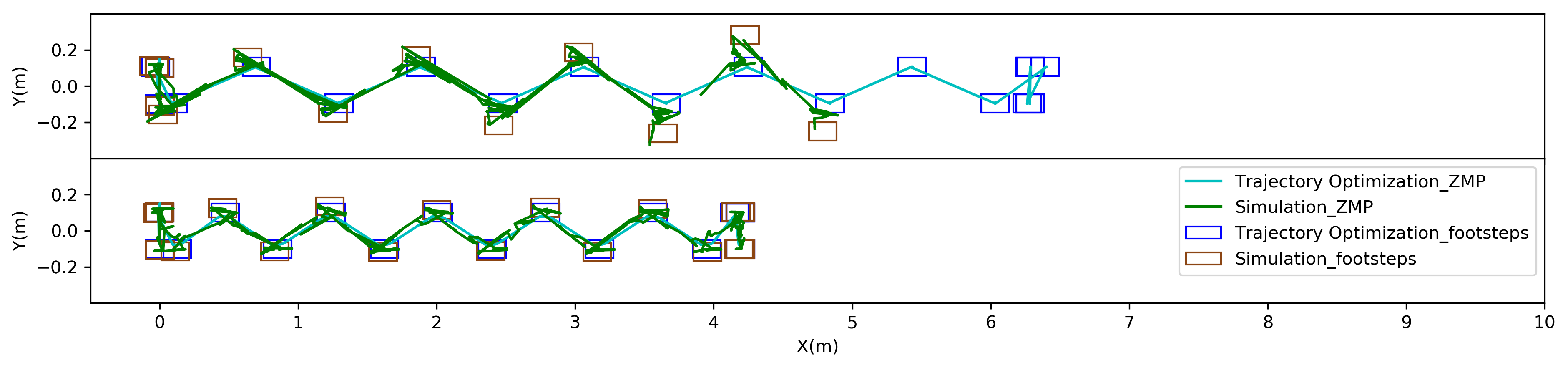}\label{footstep_slippage}}
    \caption{Scenario 1: Footstep locations \protect\subref{xvel_com_vel_tracking} with the set of weights $\alpha=1$ and $\beta=\gamma=0$, the robot succeeds to finish the walking, however the ZMP from the plan is mostly on the boundaries. \protect\subref{xvel_com_push} the robot is pushed during walking, with the set of weights in previous case it falls down (top). However, increasing $\beta$ to 70 brings the ZMP closer to the center of the foot and the robot is able to finish the task successfully (bottom). \protect\subref{xvel_com_slippage} Decreasing the friction coefficient in the simulation to 0.15 while the maximum RCoF in TO is around 0.4, the robot slips and falls down with the previous weights (top), increasing $\gamma$ to 30 decreases automatically the step length (bottom) and enables the robot to successfully complete the task.}
    \label{footstep}
    \vspace{-5 mm}
\end{figure*}

In this scenario, we show how the cost weights of the TO problem \eqref{eq:TO} affect the performance of the robot during walking in the presence of uncertainties. The desired behaviour in this scenario is to start stepping with zero walking velocity, then continue walking forward with the desired velocity of $v_{des}=1m/s$, and finally resume stepping in place at the end of motion. In the first case, we set the cost weights related to the ZMP and RCoF to zero i.e. $\beta=0, \gamma=0$ and $\alpha=1$. As a result, the CoM trajectory and footstep locations are computed using \eqref{eq:TO} under ZMP and friction cone hard constraints. Although there are several discrepancies between the simplified model and the simulation environment (dynamics model, contact model, etc.), the robot is able to achieve the task thanks to the iLQG feedback controller, as shown in Fig. \ref{xvel_com}\subref{xvel_com_vel_tracking}. However, Fig.~\ref{footstep}\subref{footstep_vel_tracking} shows that the ZMP generated by TO is on the boundaries of the support polygon, which could lead to a fall in the presence of external disturbances.

\begin{figure}[!t]
    \centering
    \includegraphics[clip,trim=0.2cm .3cm 0.2cm .2cm,width=.45\textwidth]{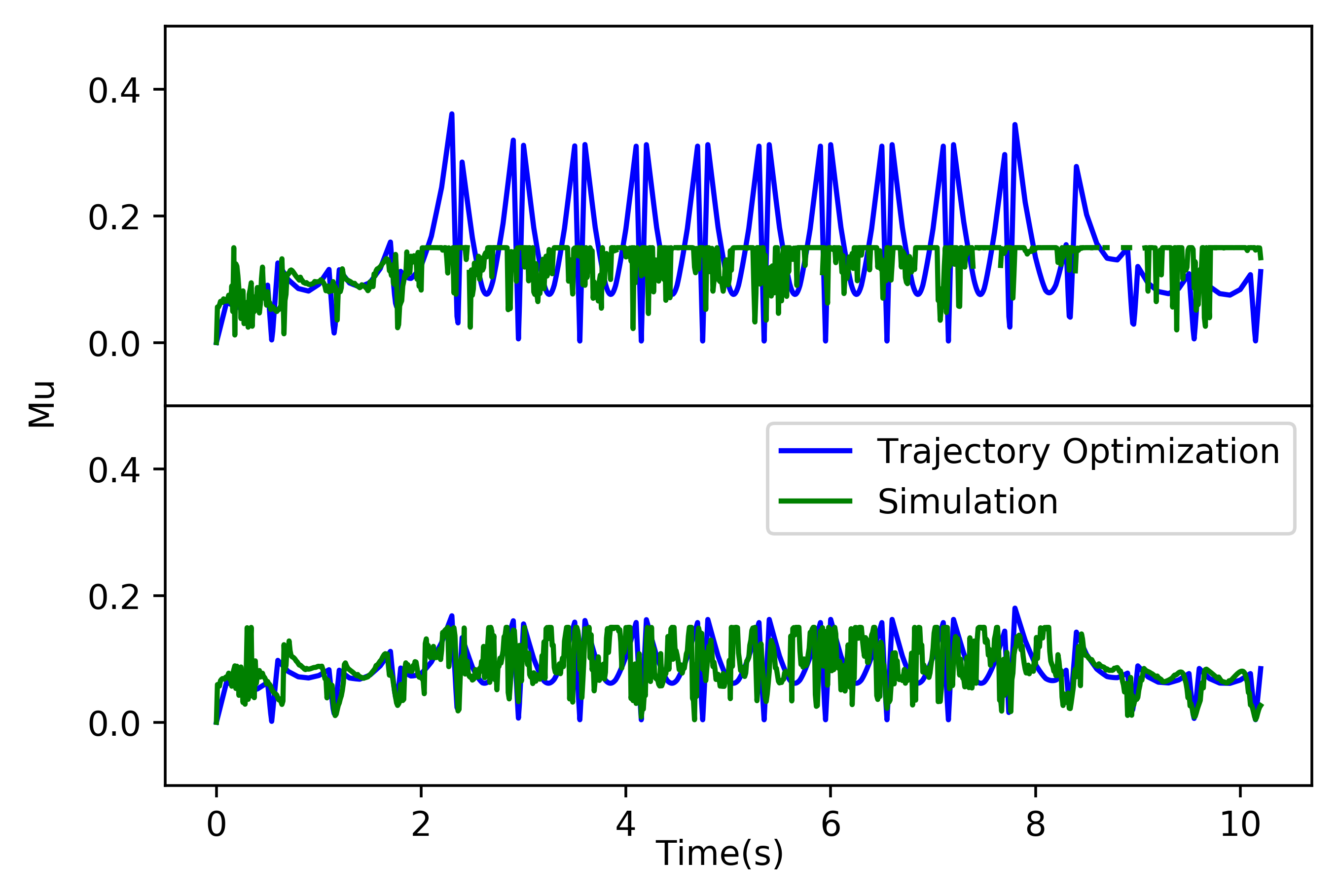}
    \caption{Scenario 1, case (c): The friction coefficient in the simulation is 0.15 while its value considered in the TO problem is 0.4. With the set of weights $\alpha=1$ and $\beta=70, \gamma=0$ the robot falls down, because the maximum RCoF is around 0.3 (top). However, increasing $\gamma$ to 30 enables the robot to successfully finish the task (bottom).}
    \label{fig:RCoF}
    \vspace{-5 mm}
\end{figure}

In the second case, we show the brittleness of the motions obtained with $\beta=0, \gamma=0$. In this simulation, we exert a lateral push $F_d=60 N$ from $ t_s = 4.9 s$ to $ t_e = 5.1 s$ to the robot, which causes a fall with $\alpha=1, \beta=0, \gamma=0$ (Fig. \ref{xvel_com}\subref{xvel_com_push}, top; see also the supplementary video). However, by setting $\beta=70$, the robot is able to successfully walk (Fig. \ref{xvel_com}\subref{xvel_com_push}, bottom). In this case, adding a high ZMP cost moves the desired ZMP trajectory from the boundaries (Fig. \ref{footstep}\subref{xvel_com_push}, top) to the middle of the foot (Fig. \ref{footstep}\subref{xvel_com_push}, bottom).

In the third case, in order to show the effect of uncertainty in the friction coefficient, we decrease the friction coefficient in the simulation to 0.15, while in the TO we considered it to be 0.4. In this case, with the same weight of the previous case, the robot loses balance and falls down (Fig. \ref{xvel_com}\subref{xvel_com_slippage}, top; see also the supplementary video), because the RCoF is higher than the real friction coefficient (Fig. \ref{fig:RCoF}, top). However, by increasing $\gamma$ to 30 the RCoF is decreased (Fig. \ref{fig:RCoF}, bottom) and the robot is able to walk without falling down (Fig. \ref{footstep}\subref{xvel_com_slippage}, bottom). This is achieved at the cost of decreasing the step length (Fig. \ref{footstep}\subref{xvel_com_slippage}, bottom), which degrades the velocity tracking (Fig. \ref{xvel_com}\subref{xvel_com_slippage}, bottom). In fact by increasing $\gamma$ the step length and walking velocity are automatically decreased to decrease the RCoF (Fig. \ref{fig:RCoF}, bottom), which enables the robot to finish the task without falling.

\subsection{Scenario 2: Using BO to find optimal cost weights of TO}
This subsection shows the application of BO (Section \ref{section:BO}) to generate robust gaits in the presence of various disturbances. We set in this scenario $\alpha=1$ to create an incentive towards viable gaits \cite{wieber2016modeling}, and optimize for $\beta$ and $\gamma$ to trade off robustness and performance. The range of weight values that we consider is $0 \leq \beta,\gamma \leq 1000$. In all cases we start with $\beta=\gamma =1000$, which leads to high ZMP and RCoF margins. We investigate four cases: without disturbances, with external pushes on the upper body, unknown decrease of the surface friction coefficient, and finally both external forces and decrease of friction coefficient.

\begin{figure*}[!t]
    \centering
    \subfloat[Walking without disturbance]{\includegraphics[clip,trim=0.2cm .3cm 0.2cm .2cm,width=.45\textwidth]{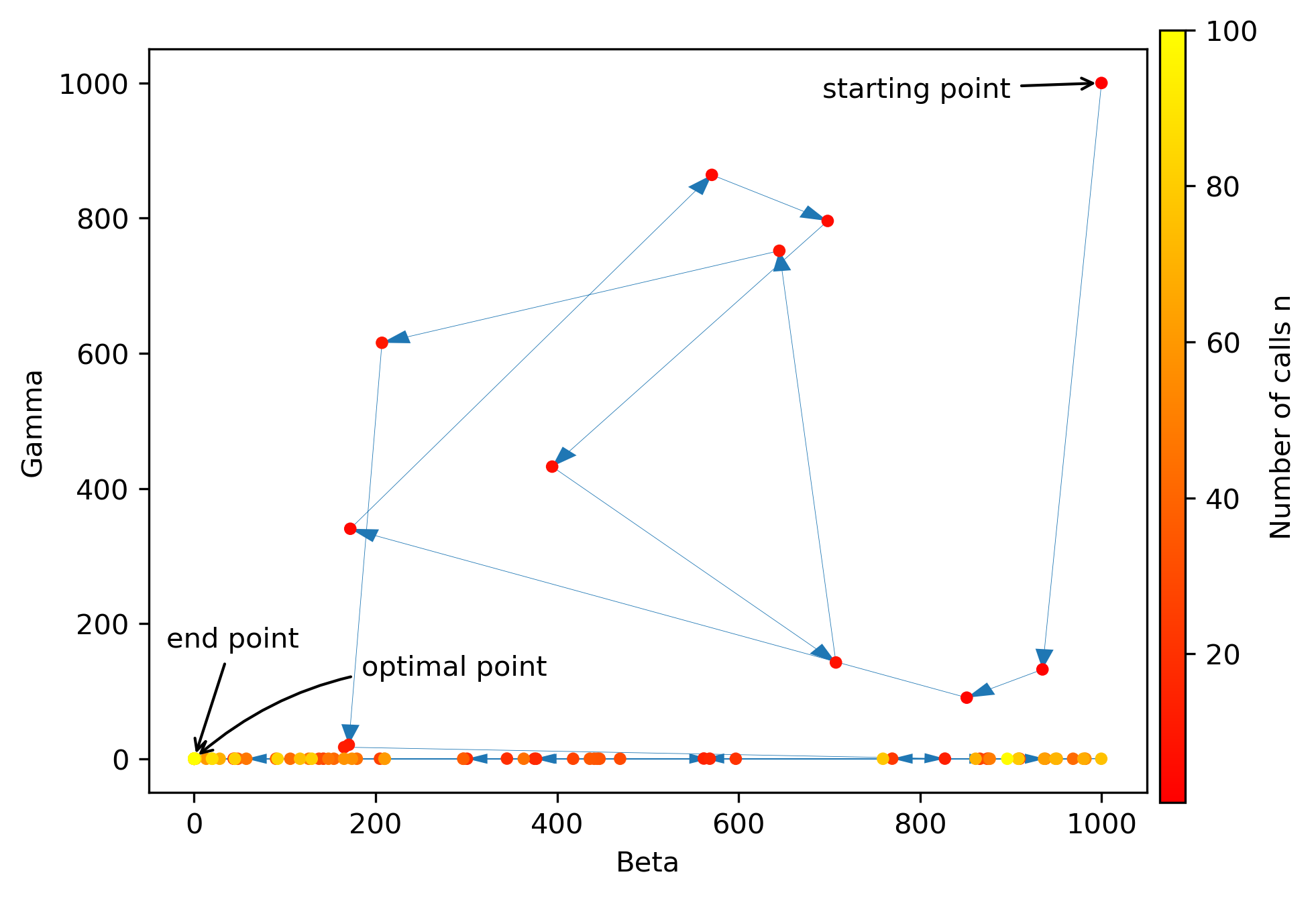}\label{nominal_weights}}
    \subfloat[Walking with external push]{\includegraphics[clip,trim=0.2cm .3cm 0.2cm .2cm,width=.45\textwidth]{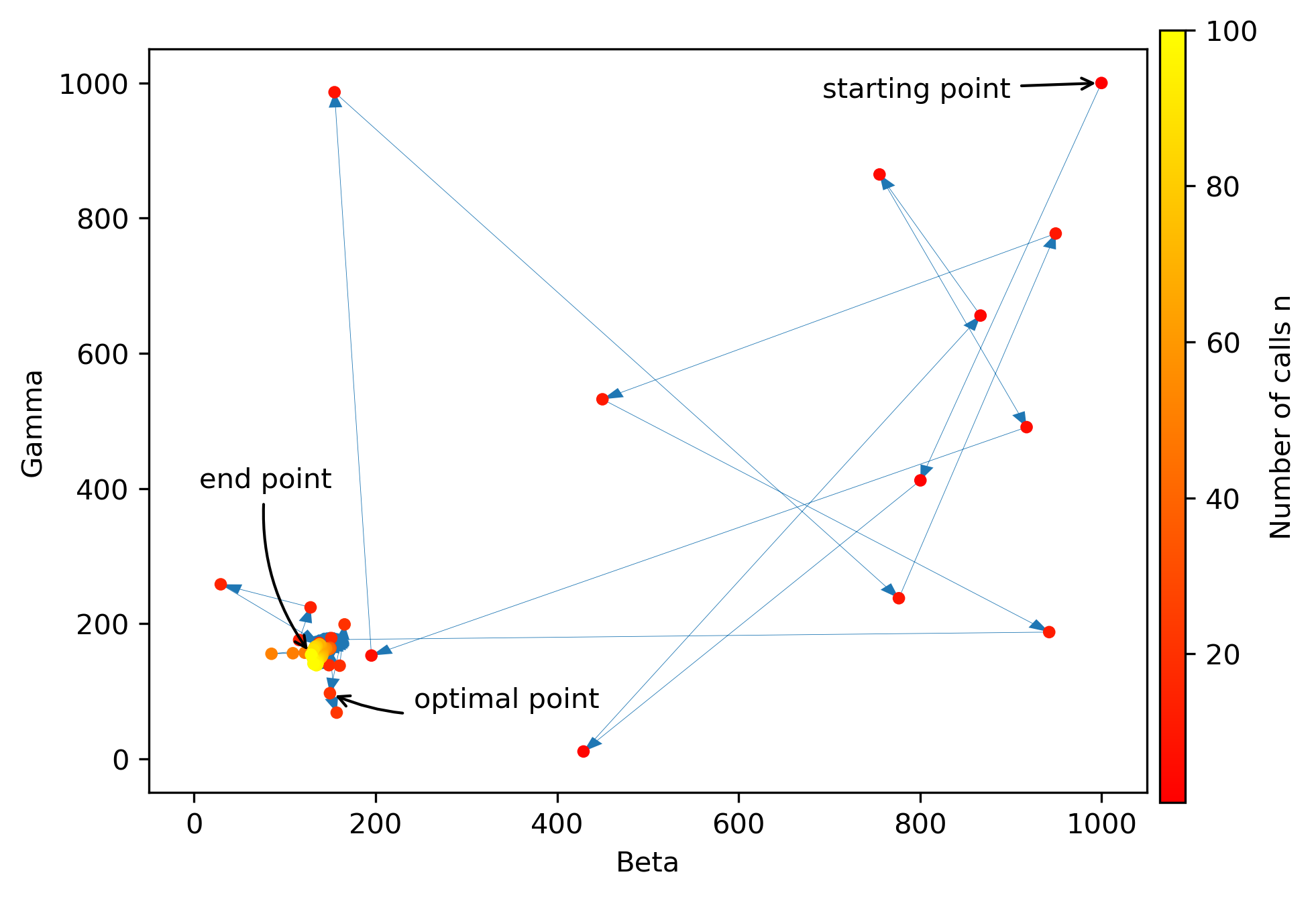}\label{force_weights}}\\
    \subfloat[Walking on slippery terrain ]{\includegraphics[clip,trim=0.2cm .3cm 0.2cm .2cm,width=.45\textwidth]{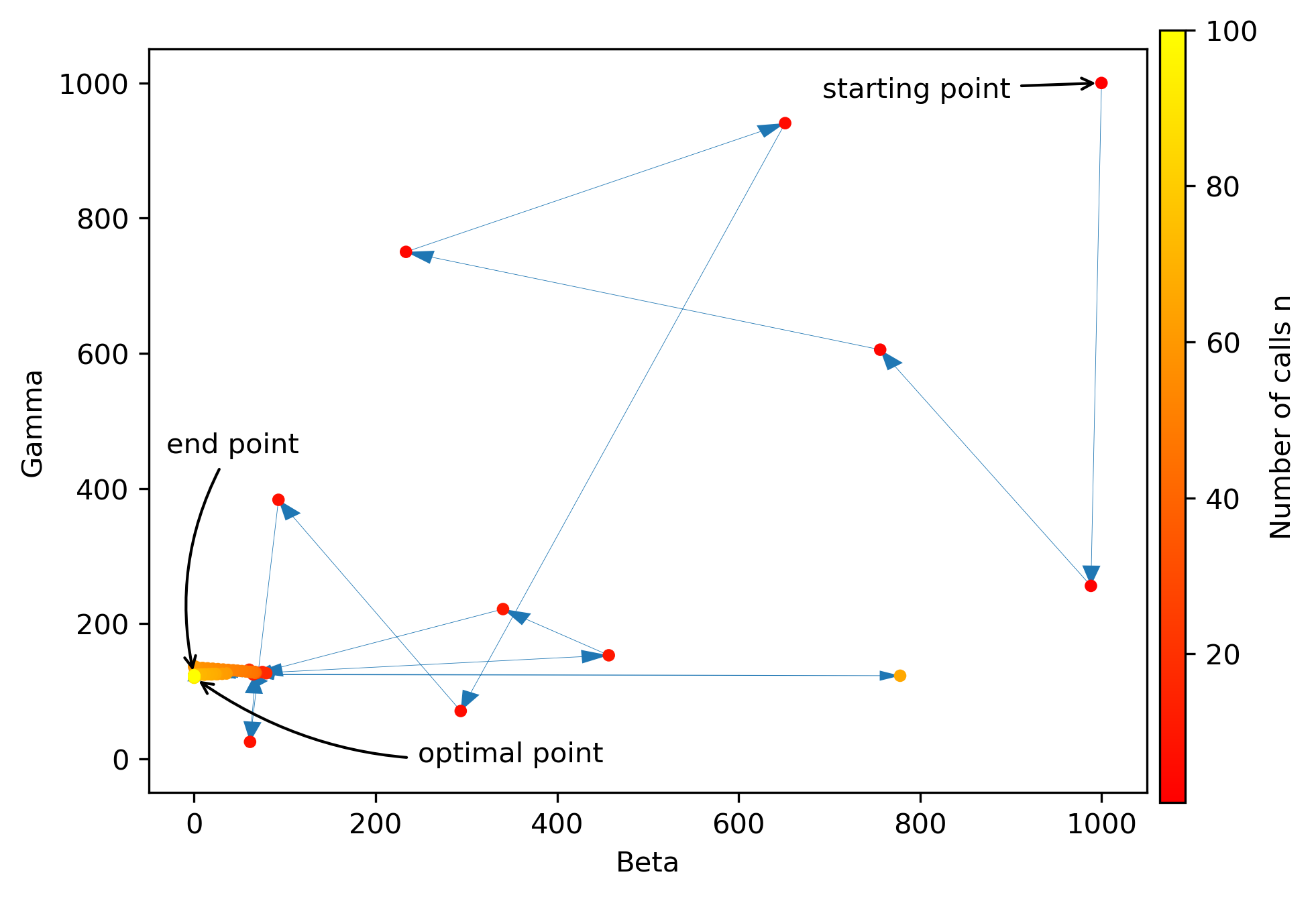}\label{slip_weights}}
    \subfloat[Walking on slippery surface with external push]{\includegraphics[clip,trim=0.2cm .3cm 0.2cm .2cm,width=.45\textwidth]{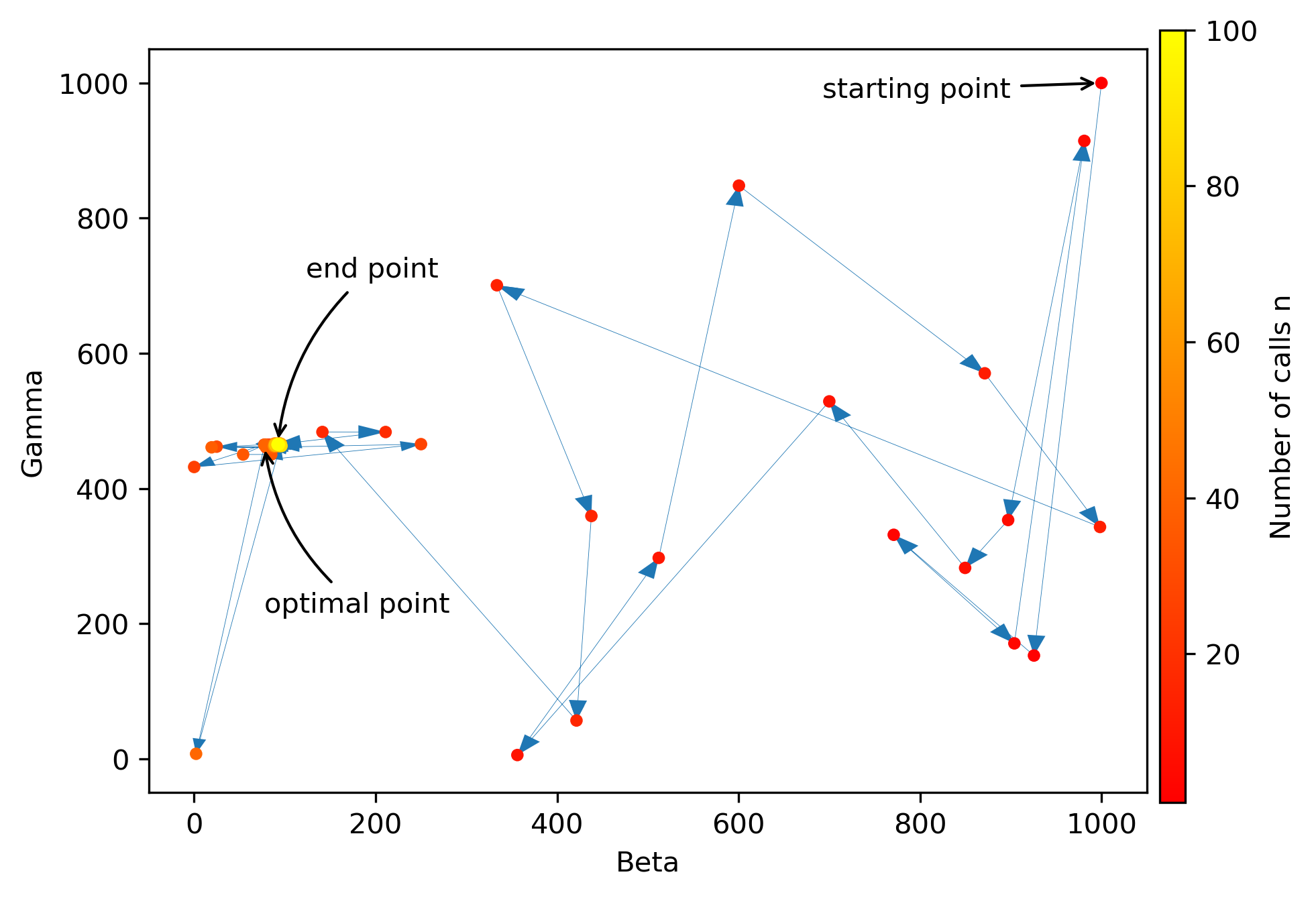}\label{slip_force_weights}}\\
    \caption{Scenario 2: History (path) of the exploration by BO starting from $\beta=\gamma=1000$ in different cases.}
    \label{weights}
    \vspace{-5 mm}
\end{figure*}

\begin{figure*}[!t]
    \centering
    \subfloat[Value of BO cost over number of calls]{\includegraphics[clip,trim=0.2cm .3cm 0.2cm .2cm,width=.45\textwidth]{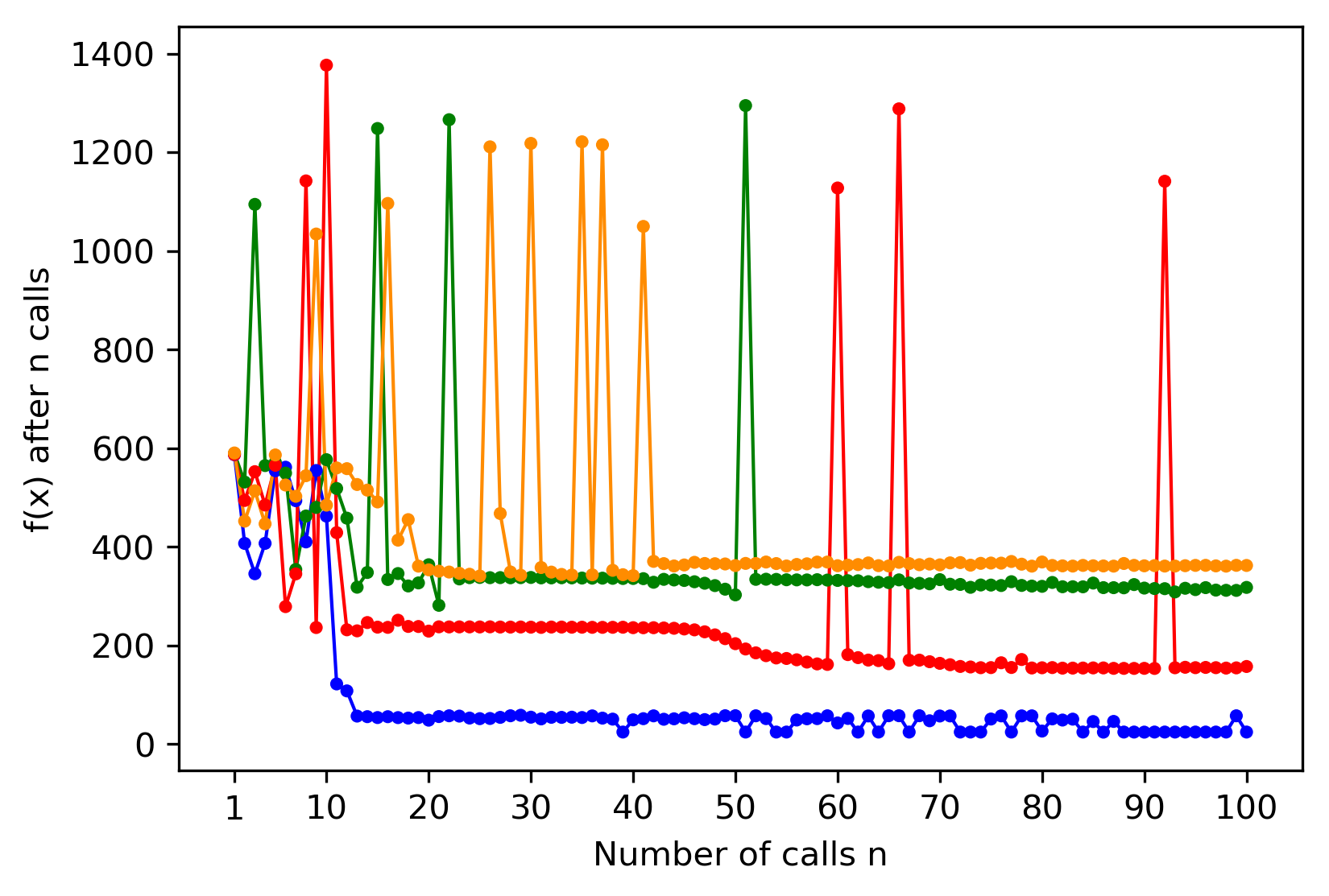}\label{original_cost}}
    \subfloat[Minimum value of cost of current call and all the last calls]{\includegraphics[clip,trim=0.cm .3cm 0.2cm .2cm,width=.45\textwidth]{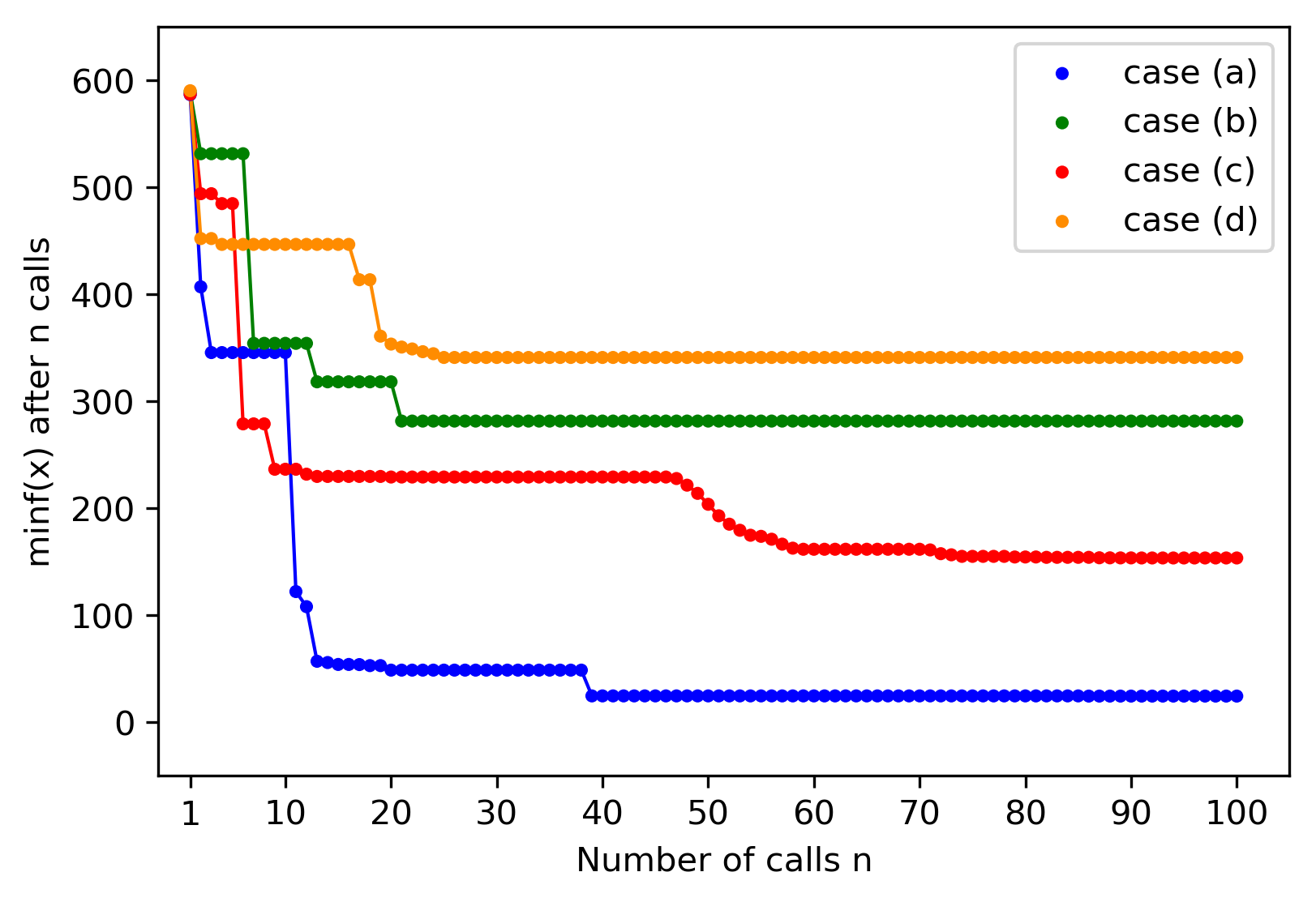}\label{minimum_cost}}
    \caption{Scenario 2: History (path) of the exploration by BO with two cost weights of TO to tune}
    \label{cost}
    \vspace{-5 mm}
\end{figure*}

In the first case, we consider no disturbance in the simulation. Since increasing $\beta,\gamma$ for having more robustness is achieved at the cost of decrease in performance (which is velocity tracking), we expect to have $\beta=0$ and $\gamma=0$ as the optimal values. As it is shown in Fig. \ref{weights}\subref{nominal_weights}, starting from $\beta=\gamma=1000$, the weights converge to zero, which confirms what we expected. Fig.~\ref{cost}\subref{minimum_cost} shows that BO converges after 39 iterations, while after 14 iteration it has already found a set of weights leading to a reasonable tracking error (small cost value).

In the second case, the robot is pushed at $t=3.1 sec$ in both forward and sideward directions with $F_d=50 N$ and $F_d=75 N$, respectively, during $\Delta t=0.2 sec$. Then at $t=6.1 sec$, the robot is pushed again sideward and backward by $F_d=65 N$ and $F_d=75 N$ during $\Delta t=0.2 s$. Note that with this set of pushes in different directions, we make sure that we cover disturbances in all directions. Also, we exert the pushes in the beginning of steps which makes it harder for the controller without step location and timing adaptation to recover from it \cite{khadiv2018walking}. We apply BO to find the optimal weights for this case. The points evaluated by BO are shown in Fig.~\ref{weights}\subref{force_weights}. The optimal point is $\beta=149.63, \gamma=96.94$ which is obtained after 21 iterations. The weight on the ZMP ($\beta$) is increased to bring the ZMP closer to the center during walking, enabling the robot to reject disturbances while minimizing the performance error (walking with a desired velocity).

In the third case, we test the robot walking on a surface with unexpected drop on the friction coefficient. In this case we decrease the friction coefficient of the surface to 0.1, while the value set for the TO constraint is 0.4. 
The optimal value of the weights obtained from the BO (after 75 calls, see Fig. \ref{cost}\subref{minimum_cost}) is  $\beta=0, \gamma=120.05$. In this case, the optimal weights penalize more the RCoF so that the planned motion requires less friction to be realized. As shown in Fig. \ref{cost}\subref{minimum_cost}, after 9 iterations BO has already found weights leading to a decent cost.

In the final case, we exert the external push (as in the second case) and drop the friction coefficient to 0.15. Starting again from $\beta=\gamma=1000$, the weights converge to $\beta=77.23, \gamma=426.83$ (Fig. \ref{weights}\subref{slip_force_weights}) after 25 calls of BO (see Fig. \ref{cost}\subref{minimum_cost}). In this case, both weights are increased to make the gait robust to the exerted disturbances, while at the same time these weights yield the best task achievement (walking velocity tracking).

Fig.~\ref{cost}\subref{original_cost} shows the evolution of cost values of BO over iterations for all cases of this scenario. The high jumps on the value are due to the high penalty given to falls. As expected, the decrease in the cost is not monotonic, because we are not using gradient-based optimization. Fig.~\ref{cost}\subref{minimum_cost} plots the minimum value of the current cost and all last calls for evaluating the function. The interesting point is that for all the cases after a few iterations (less than 20 calls), the cost has already settled. This suggests that we do not need to do many experiments, which can be costly (especially for humanoid robots) and time consuming.

\subsection{Scenario 3: Using BO to tune more cost parameters}

\begin{figure}[!t]
    \centering
    \includegraphics[clip,trim=0.2cm .3cm 0.2cm .2cm,width=.48\textwidth]{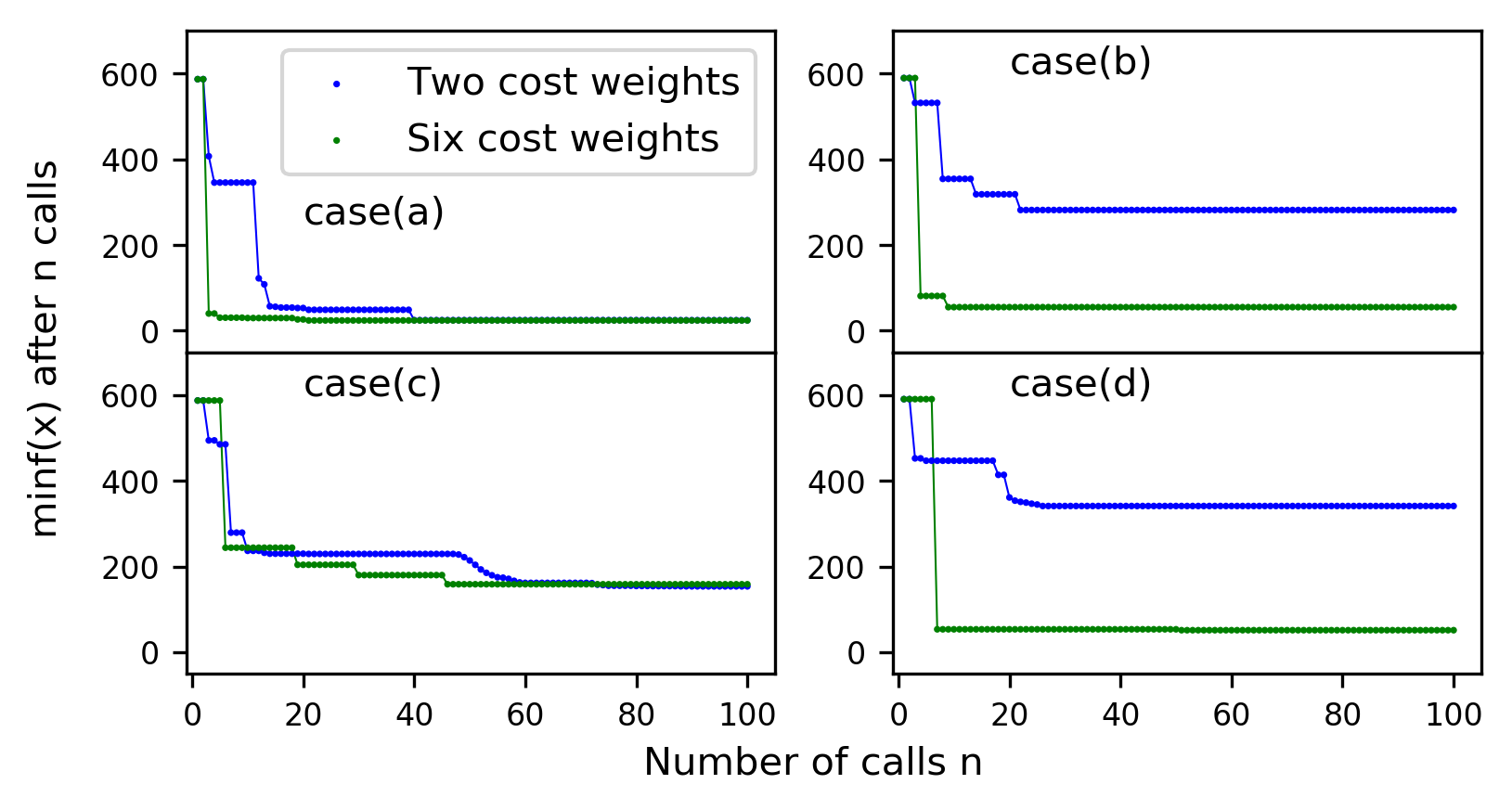}
    \caption{Scenario 3: The minimum value of cost of current call and all the last calls for the cases with two and six cost weights.}
    \label{cost_3}
    \vspace{-5 mm}
\end{figure}

In this scenario, we investigate the use of BO for higher dimensional problems to demonstrate that it can scale to more
complex cost functions. To do so, we use the same TO problem in \eqref{eq:TO}, but with different cost weights in the lateral and sagittal directions. In this case, we have six weights to tune, i.e. $(\alpha_{x,y},\beta_{x,y},\gamma_{x,y})$. The ranges of weight values that we consider are $1 \leq \alpha_{x,y} \leq 100$ and $0 \leq \beta_{x,y},\gamma_{x,y} \leq 1000$. In all cases we start with $\alpha_{x,y}=1$ and $\beta_{x,y}=\gamma_{x,y} =1000$. As we see in Fig. \ref{cost_3}, BO is able to find a good solution after a few iterations, despite the larger number of parameters to tune. 
In this case, BO has more degrees of freedom to take different TO cost values for sagittal and lateral directions. The final cost values reveal that in this case, a lower value of the BO cost was found (apart from the nominal case where both converged to the same cost). This result suggests that our approach can scale to larger problems. 


\section{CONCLUSIONS and future work}\label{section:conclusion}

In this paper, we presented an approach combining gradient-based trajectory optimization with black-box data-driven optimization to achieve robust gaits for humanoid locomotion. We have used Bayesian optimization to find the best cost weights of the trajectory optimization problem for full robot walking with different disturbances. Our simulation results showed that this approach can find the best cost weights that trade off robustness against performance with only a few evaluations.

As we seek robust solutions to an optimization problem, a natural question arises: what is the connection between our method and \emph{robust optimization (RO)}? In particular, the recent data-driven RO~\cite{bertsimas2018data} that also identifies uncertainty sets from data. A typical robust constraint can be formulated as $f_0(x, w) \leq 0, \forall w\in W$, e.g. $x+w\leq 0, \forall w \in W$. However, it is hard to write down constraints in such an explicit form for the overall closed-loop system in Fig.~\ref{fig:block_diagram} with respect to the hyper-parameter $\delta$. Hence we resort here to a black-box scheme such as BO with the limited modeling insight we have. We are currently investigating the connection between our approach and data-driven RO \cite{bertsimas2018data} in this context.


An interesting extension of this work, which is of great value for implementation on real hardware, is the safety of exploration. Another interesting extension is to use simulation data to modify simultaneously the iLQG and TO cost weights.
Another future direction is to use \eqref{eq:TO} in an MPC fashion. There will be a lot of interesting problems to tackle in this case, such as recursive feasibility \cite{villa2017model}, terminal cost, etc. Also, we are interested in using the same approach for more complex models in trajectory optimization, e.g. centroidal dynamics \cite{ponton2018time} or full robot dynamics \cite{budhiraja2018differential}.

\bibliography{Master}

\begin{thebibliography}{10}
\providecommand{\url}[1]{#1}
\csname url@rmstyle\endcsname
\providecommand{\newblock}{\relax}
\providecommand{\bibinfo}[2]{#2}
\providecommand\BIBentrySTDinterwordspacing{\spaceskip=0pt\relax}
\providecommand\BIBentryALTinterwordstretchfactor{4}
\providecommand\BIBentryALTinterwordspacing{\spaceskip=\fontdimen2\font plus
\BIBentryALTinterwordstretchfactor\fontdimen3\font minus
  \fontdimen4\font\relax}
\providecommand\BIBforeignlanguage[2]{{%
\expandafter\ifx\csname l@#1\endcsname\relax
\typeout{** WARNING: IEEEtran.bst: No hyphenation pattern has been}%
\typeout{** loaded for the language `#1'. Using the pattern for}%
\typeout{** the default language instead.}%
\else
\language=\csname l@#1\endcsname
\fi
#2}}

\bibitem{kajita2003biped}
S.~Kajita, F.~Kanehiro, K.~Kaneko, K.~Fujiwara, K.~Harada, K.~Yokoi, and
  H.~Hirukawa, ``Biped walking pattern generation by using preview control of
  zero-moment point,'' in \emph{Robotics and Automation (ICRA), IEEE
  International Conference on}.\hskip 1em plus 0.5em minus 0.4em\relax IEEE,
  2003, pp. 1620--1626.

\bibitem{kajita20013d}
S.~Kajita, F.~Kanehiro, K.~Kaneko, K.~Yokoi, and H.~Hirukawa, ``The 3d linear
  inverted pendulum mode: A simple modeling for a biped walking pattern
  generation,'' in \emph{Intelligent Robots and Systems,IEEE/RSJ International
  Conference on}.\hskip 1em plus 0.5em minus 0.4em\relax IEEE, 2001, pp.
  239--246.

\bibitem{wieber2006trajectory}
P.~B. Wieber, ``Trajectory free linear model predictive control for stable
  walking in the presence of strong perturbations,'' in \emph{2006 6th IEEE-RAS
  International Conference on Humanoid Robots}.\hskip 1em plus 0.5em minus
  0.4em\relax IEEE, 2006, pp. 137--142.

\bibitem{herdt2010online}
A.~Herdt, H.~Diedam, P.-B. Wieber, D.~Dimitrov, K.~Mombaur, and M.~Diehl,
  ``Online walking motion generation with automatic footstep placement,''
  \emph{Advanced Robotics}, vol.~24, no. 5-6, pp. 719--737, 2010.

\bibitem{khadiv2017pattern}
M.~Khadiv, S.~A.~A. Moosavian, A.~Herzog, and L.~Righetti, ``Pattern generation
  for walking on slippery terrains,'' in \emph{2017 5th RSI International
  Conference on Robotics and Mechatronics (ICRoM)}.\hskip 1em plus 0.5em minus
  0.4em\relax IEEE, 2017, pp. 120--125.

\bibitem{orin2013centroidal}
D.~E. Orin, A.~Goswami, and S.-H. Lee, ``Centroidal dynamics of a humanoid
  robot,'' \emph{Autonomous Robots}, vol.~35, no. 2-3, pp. 161--176, 2013.

\bibitem{wieber2016modeling}
P.-B. Wieber, R.~Tedrake, and S.~Kuindersma, ``Modeling and control of legged
  robots,'' in \emph{Springer Handbook of Robotics}.\hskip 1em plus 0.5em minus
  0.4em\relax Springer, 2016, pp. 1203--1234.

\bibitem{herzog2015trajectory}
A.~Herzog, N.~Rotella, S.~Schaal, and L.~Righetti, ``Trajectory generation for
  multi-contact momentum control,'' in \emph{Humanoid Robots (Humanoids), 2015
  IEEE-RAS 15th International Conference on}.\hskip 1em plus 0.5em minus
  0.4em\relax IEEE, 2015, pp. 874--880.

\bibitem{carpentier2016A}
J.~Carpentier, S.~Tonneau, M.~Naveau, O.~Stasse, and N.~Mansard, ``{A Versatile
  and Efficient Pattern Generator for Generalized Legged Locomotion},'' in
  \emph{{IEEE International Conference on Robotics and Automation (ICRA)}},
  Stockholm, Sweden, May 2016.

\bibitem{ponton2018time}
B.~Ponton, A.~Herzog, A.~Del~Prete, S.~Schaal, and L.~Righetti, ``On time
  optimization of centroidal momentum dynamics,'' in \emph{2018 IEEE
  International Conference on Robotics and Automation (ICRA)}.\hskip 1em plus
  0.5em minus 0.4em\relax IEEE, 2018, pp. 1--7.

\bibitem{carpentier2017learning}
J.~Carpentier, R.~Budhiraja, and N.~Mansard, ``Learning feasibility constraints
  for multi-contact locomotion of legged robots,'' in \emph{Robotics: Science
  and Systems}, 2017, p.~9p.

\bibitem{lengagne2013generation}
S.~Lengagne, J.~Vaillant, E.~Yoshida, and A.~Kheddar, ``Generation of
  whole-body optimal dynamic multi-contact motions,'' \emph{The International
  Journal of Robotics Research}, vol.~32, no. 9-10, pp. 1104--1119, 2013.

\bibitem{shahriari2016taking}
B.~Shahriari, K.~Swersky, Z.~Wang, R.~P. Adams, and N.~De~Freitas, ``Taking the
  human out of the loop: A review of bayesian optimization,'' \emph{Proceedings
  of the IEEE}, vol. 104, no.~1, pp. 148--175, 2016.

\bibitem{lizotte2007automatic}
D.~J. Lizotte, T.~Wang, M.~H. Bowling, and D.~Schuurmans, ``Automatic gait
  optimization with gaussian process regression.'' in \emph{IJCAI}, vol.~7,
  2007, pp. 944--949.

\bibitem{cully2015robots}
A.~Cully, J.~Clune, D.~Tarapore, and J.-B. Mouret, ``Robots that can adapt like
  animals,'' \emph{Nature}, vol. 521, no. 7553, p. 503, 2015.

\bibitem{marco2016automatic}
A.~Marco, P.~Hennig, J.~Bohg, S.~Schaal, and S.~Trimpe, ``Automatic lqr tuning
  based on gaussian process global optimization,'' in \emph{2016 IEEE
  international conference on robotics and automation (ICRA)}.\hskip 1em plus
  0.5em minus 0.4em\relax IEEE, 2016, pp. 270--277.

\bibitem{su2018sample}
Y.~Su, Y.~Wang, and A.~Kheddar, ``Sample-efficient learning of soft task
  priorities through bayesian optimization,'' in \emph{2018 IEEE-RAS 18th
  International Conference on Humanoid Robots (Humanoids)}.\hskip 1em plus
  0.5em minus 0.4em\relax IEEE, 2018, pp. 1--6.

\bibitem{seyde2019locomotion}
T.~Seyde, J.~Carius, R.~Grandia, F.~Farshidian, and M.~Hutter, ``Locomotion
  planning through a hybrid bayesian trajectory optimization,'' \emph{arXiv
  preprint arXiv:1903.03823}, 2019.

\bibitem{calandra2014experimental}
R.~Calandra, A.~Seyfarth, J.~Peters, and M.~P. Deisenroth, ``An experimental
  comparison of bayesian optimization for bipedal locomotion,'' in \emph{2014
  IEEE International Conference on Robotics and Automation (ICRA)}.\hskip 1em
  plus 0.5em minus 0.4em\relax IEEE, 2014, pp. 1951--1958.

\bibitem{antonova2016sample}
R.~Antonova, A.~Rai, and C.~G. Atkeson, ``Sample efficient optimization for
  learning controllers for bipedal locomotion,'' in \emph{2016 IEEE-RAS 16th
  International Conference on Humanoid Robots (Humanoids)}.\hskip 1em plus
  0.5em minus 0.4em\relax IEEE, 2016, pp. 22--28.

\bibitem{rai2018bayesian}
A.~Rai, R.~Antonova, S.~Song, W.~Martin, H.~Geyer, and C.~Atkeson, ``Bayesian
  optimization using domain knowledge on the atrias biped,'' in \emph{2018 IEEE
  International Conference on Robotics and Automation (ICRA)}.\hskip 1em plus
  0.5em minus 0.4em\relax IEEE, 2018, pp. 1771--1778.

\bibitem{charbonneau2018learning}
M.~Charbonneau, V.~Modugno, F.~Nori, G.~Oriolo, D.~Pucci, and S.~Ivaldi,
  ``Learning robust task priorities of qp-based whole-body
  torque-controllers,'' in \emph{2018 IEEE-RAS 18th International Conference on
  Humanoid Robots (Humanoids)}.\hskip 1em plus 0.5em minus 0.4em\relax IEEE,
  2018, pp. 1--9.

\bibitem{yuan2019bayesian}
K.~Yuan, I.~Chatzinikolaidis, and Z.~Li, ``Bayesian optimization for whole-body
  control of high-degree-of-freedom robots through reduction of
  dimensionality,'' \emph{IEEE Robotics and Automation Letters}, vol.~4, no.~3,
  pp. 2268--2275, 2019.

\bibitem{tassa2012synthesis}
Y.~Tassa, T.~Erez, and E.~Todorov, ``Synthesis and stabilization of complex
  behaviors through online trajectory optimization,'' in \emph{Intelligent
  Robots and Systems (IROS), 2012 IEEE/RSJ International Conference on}.\hskip
  1em plus 0.5em minus 0.4em\relax IEEE, 2012, pp. 4906--4913.

\bibitem{tassa2014control}
Y.~Tassa, N.~Mansard, and E.~Todorov, ``Control-limited differential dynamic
  programming,'' in \emph{Robotics and Automation (ICRA), 2014 IEEE
  International Conference on}.\hskip 1em plus 0.5em minus 0.4em\relax IEEE,
  2014, pp. 1168--1175.

\bibitem{erez2013integrated}
T.~Erez, K.~Lowrey, Y.~Tassa, V.~Kumar, S.~Kolev, and E.~Todorov, ``An
  integrated system for real-time model predictive control of humanoid
  robots,'' in \emph{2013 13th IEEE-RAS International Conference on Humanoid
  Robots (Humanoids)}.\hskip 1em plus 0.5em minus 0.4em\relax IEEE, 2013, pp.
  292--299.

\bibitem{brochu2010tutorial}
E.~Brochu, V.~M. Cora, and N.~De~Freitas, ``A tutorial on bayesian optimization
  of expensive cost functions, with application to active user modeling and
  hierarchical reinforcement learning,'' \emph{arXiv preprint arXiv:1012.2599},
  2010.

\bibitem{hoffman2011portfolio}
M.~D. Hoffman, E.~Brochu, and N.~de~Freitas, ``Portfolio allocation for
  bayesian optimization.'' in \emph{UAI}.\hskip 1em plus 0.5em minus
  0.4em\relax Citeseer, 2011, pp. 327--336.

\bibitem{khadiv2018walking}
M.~Khadiv, A.~Herzog, S.~A.~A. Moosavian, and L.~Righetti, ``Walking control
  based on step timing adaptation,'' \emph{arXiv preprint arXiv:1704.01271v2},
  2018.

\bibitem{bertsimas2018data}
D.~Bertsimas, V.~Gupta, and N.~Kallus, ``Data-driven robust optimization,''
  \emph{Mathematical Programming}, vol. 167, no.~2, pp. 235--292, 2018.

\bibitem{villa2017model}
N.~A. Villa and P.-B. Wieber, ``Model predictive control of biped walking with
  bounded uncertainties,'' in \emph{2017 IEEE-RAS 17th International Conference
  on Humanoid Robotics (Humanoids)}.\hskip 1em plus 0.5em minus 0.4em\relax
  IEEE, 2017, pp. 836--841.

\bibitem{budhiraja2018differential}
R.~Budhiraja, J.~Carpentier, C.~Mastalli, and N.~Mansard, ``Differential
  dynamic programming for multi-phase rigid contact dynamics,'' in \emph{2018
  IEEE-RAS 18th International Conference on Humanoid Robots (Humanoids)}.\hskip
  1em plus 0.5em minus 0.4em\relax IEEE, 2018, pp. 1--9.

\end{thebibliography}
\bibliographystyle{IEEEtran}

\end{document}